# Waymo's Fatigue Risk Management Framework: Prevention, Monitoring, and Mitigation of Fatigue-Induced Risks while Testing Automated Driving Systems

Francesca Favaro', Keith Hutchings,
Philip Nemec, Leticia Cavalcante, Trent Victor[1]

## Abstract

Fatigue is a recognized contributory factor in a substantial fraction of on-road crashes involving human drivers, and mitigation of fatigue-induced risks is still an open concern researched world-wide. Risks that human drivers are inherently subject to are one of the very reasons why Waymo committed to Automated Driving System (ADS) technology in the first place. Yet, responsible development of this technology requires developers to investigate and address fatigue-induced risks as they pertain to testing operations.

This report thus presents Waymo's proposal for a systematic fatigue risk management framework that addresses prevention, monitoring, and mitigation of fatigue-induced risks during on-road testing of ADS technology. The proposed framework remains flexible to incorporate continuous improvements, and was informed by state of the art practices, research, learnings, and experience (both internal and external to our company).

Today, Waymo operates a fully autonomous commercial ride-hailing service - Waymo One[2] - in multiple locations, including Arizona and California. Waymo One showcases Waymo's operations of its SAE Level 4 technology[3], which does not rely on human intervention. Yet, human operators - called autonomous specialists at Waymo - play an important role in helping to ensure on-road safety during development and testing. Unlike drivers of traditional vehicles, autonomous specialists do not execute navigation, path planning, or control tasks during fully autonomous operation testing missions. Instead, they are highly trained to focus on overseeing the Waymo Driver's operations. Waymo puts a strong emphasis on the need for continued training, countermeasures, assessment of autonomous specialists, as well as tracking of operational performance and procedures. The study of a substantial body of literature on human driver inattention, and observations from our own early testing have led us to believe that risks involved with human driving can be most effectively addressed through technology that handles the entire driving task. Testing of our technology (when a human being is still involved) involves the diligent review and improvement of evolving fatigue mitigation techniques, some of which were developed from the ground up for Waymo's testing program.

---

[1] The authors indicated are the main drafters of the present publication. Many others across Waymo contributed to the creation and implementation of the Fatigue Risk Management framework here presented.

[2] https://waymo.com/waymo-one/

[3] The scope of this paper does not apply to testing of Waymo's commercial motor vehicles, which continues to be informed by and adopts many aspects of this FRM program, but also involves separate regulatory requirements (e.g., 49 CFR Part 395) related to the fatigue of autonomous specialists. Motor carriers and drivers of commercial motor vehicles also have the benefit of the North American Fatigue Management Program, which offers information and training resources for fatigue management.

Waymo's proposed Fatigue Risk Management (FRM) framework is informed by the substantial literature and operational experience of multiple industries: both in the transportation sector (e.g., aviation, railways, commercial motor vehicles, and more recent practices established within the AV industry), as well as non-transportation related sectors (e.g., nuclear and chemical). It follows the principles and processes of a Safety Management System whereby specific hazards are identified and risk is managed systematically. Waymo's FRM framework revolves around three major building blocks of: (1) fatigue prevention; (2) fatigue monitoring; and (3) fatigue mitigation. Rather than in isolation, each block encompasses methodologies that feed the FRM cycle at multiple stages, and that span the totality of testing operations before, while, and after driving. Specifically, the practical methodologies that make up this framework are organized under the headings of: (1) continued education; (2) awareness and reporting; (3) real-time vigilance assessment; (4) supplemental engagement; and (5) adaptive scheduling. Together, these elements feed into a holistic framework for countering fatigue-induced risks, for a total of twenty countermeasures. Throughout over ten years of on-road operations, Waymo's learnings showcase how the framework presented in this paper provides a valid and useful structure to effectively prioritize and address fatigue countermeasures. Waymo has gone to great lengths to develop this framework by engaging with experts in fatigue, leveraging fatigue monitoring technology, and building unique processes and technology to mitigate the potential for fatigue-related incidents during our testing operations.

**Table of Contents**

## Acronyms

| AAA | American Automobile Association |
| --- | --- |
| ADS | Automated Driving System |
| DDT | Dynamic Driving Task |
| DMS | Driver Monitoring System |
| FRM | Fatigue Risk Management |
| HMI | Human Machine Interface |
| ICT | Interactive Cognitive (In-Car) Task |
| OEDR | Object and Event Detection and Response |
| ORD | Observer Rating of Drowsiness |
| NHTSA | National Highway Traffic Safety Administration |
| PFS | Periodic Fatigue Survey |
| RT | Real-time |
| SA | Secondary Alert |
| SMS | Safety Management System |
| VTTI | Virginia Tech Transportation Institute |

# 1. Introduction and Background

## ORIGINS OF COMPLACENCY CHALLENGES

Waymo began as the Google Self-Driving Car Project in 2009. Early testing in 2012 and 2013 gave Google employees (who were not involved with the system development) the opportunity to sit behind the wheel and experience the "ancestor" of what the Waymo Driver is today - a system that, back then, was designed to control the in-lane dynamic driving task in freeway environments, and that required continuous monitoring by, and potential fallback to, the human driver. The short-lived internal testing program asked Google employees to oversee the continued operation of the system, following a 2-hour training on the system's capabilities and limitations. While employees were aware of being monitored during the testing program, the in-vehicle footage revealed that they still engaged in a number of less than ideal behaviors, including distractions, and were subject to signs of fatigue[4]. Microsleep events were observed, and some participants were also seen asleep at the wheel for extended periods of time. We further observed some participants' extensive use of personal devices and engagement in tasks that involved drivers taking their hands off the steering wheel for prolonged durations - all indications of over-reliance and complacency in a system that was explicitly stated as being experimental and requiring human monitoring. Furthermore, we observed the frequency of these behaviors increase with the length of the driving session[5]. By the end of this testing program, every participant in the program had demonstrated clear instances of inattention.

Over the past 50 years, extensive research on human operator performance in automated systems across multiple industries has indicated that human factors concerns are a difficult hurdle to overcome during prolonged monitoring of automated functionalities [1-11]. A main conclusion from the human factors literature on human performance while operating automated systems [12-15] and from human factors guidelines [4; 16] is that a general relationship between an *increasing degree* of automation and a corresponding *reduction* in human performance exists. This general phenomenon has been termed the "*irony of automation*" [1]. In other words, and in the context of driving, the better the automation, the less attention human drivers will pay to traffic and the system, and the less capable they will be to resume control [10; 17; 18]. This phenomenon is common across varying automation domains, such as process control (e.g., chemical and nuclear) and aviation (see, for example, a meta-analysis of 18 studies in [13]), and remains to date an active area of research. While the high-level paradox of smarter technologies thwarting timely and effective operator interventions remains intuitive and has a general consensus, more research is needed to understand and quantify its precise effects in *driving automation*. For instance, more research is needed on how extended periods of monitoring automated operations affect degradation of driver response to automation limitations [11; 19; 20]. Agreement is established on the heightened difficulty for humans to monitor automation effectively and to be suddenly ready to solve critical issues and regain appropriate situational awareness, once being allowed to remain out of the loop for extended periods of time. Within the existing literature, this has been largely attributed to the operators' tendency to decrease monitoring levels for highly reliable automation due to the ability to function properly for an extended period (see for instance, [5; 21]).

---

[4] See, "Why we're pursuing full autonomy": https://www.youtube.com/watch?v=6ePWBBrWSzo&t=17s
[5] Driving sessions that exceeded 30 minutes were 6 times more likely to contain incautious behavior compared to sessions under 15 minutes (66% vs 11%, respectively). These effects were evident within two weeks of continued use and are indicative of the risks of complacency and potential misuse.

This phenomenon partly explains the behavior of the participants that we saw in our earlier 2012/2013 testing, discussed before, and mitigating against this is one of the reasons why we decided to redirect our efforts toward developing Level 4 automation[6], a solution that does not rely on a human driver for any reason, and that provides the best option for achieving enhanced safety through automation [22].

As we build the World's Most Experienced Driver™, autonomous specialists - who oversee testing missions, as described later in this paper -  play an important role in the safe development of an ADS. Our early experience in 2012 and 2013 revealed the need for meticulous evaluation of the design of interactions between a human driver and the vehicle's systems when safety depends on the joint collaborations between them. It also revealed that proper training and resources to counteract fatigue-induced risks and automation complacency need not only to be designed and soundly implemented, but also continuously reassessed and evaluated. To that end, Waymo puts a strong emphasis on fatigue risk management (FRM), encompassing preventative measures that reduce the likelihood of fatigue, as well as real-time monitoring and timely mitigation of potential fatigue-related events. To date, drowsiness and fatigue remain complex phenomena with open challenges in both their scientific understanding and in the efficacy and comparison across different countermeasures, which Waymo continues to investigate in-house with a team of world-class UX and safety researchers. We iterate on our methods as necessary, but found that the overarching framework and safety strategy presented in this paper can provide value to others that are today facing the same challenges we are, with the common goals of improving road safety. At the same time, it is important to understand that solutions also need to be catered to specific use-cases and can be dependent on the maturity and stage of the individual ADS developer.

---

[6] Waymo's automated driving system is an SAE L4 system and, as such, does not depend on human monitoring or intervention for safe performance. However, in testing this system, Waymo often operates with an autonomous specialist who is employed by a transportation partner and trained to take over control in certain circumstances. This helps ensure safety as the ADS progresses to the stage where it is ready for fully autonomous operation. SAE J3016 makes clear that an ADS designed to be L4 is in fact an L4 ADS even if an autonomous specialist is assigned to monitor and intervene during its testing [23].

On-road testing missions at Waymo are varied, and may include both periods of overseeing the Waymo Driver during fully autonomous operation and periods of manual driving. Missions can, for example, be tied to generating maps, undergoing durability testing, testing new hardware, and testing new software. Unlike drivers of traditional vehicles, autonomous specialists do not execute navigation, path planning, or control tasks during fully autonomous operation testing missions. Instead, they focus on overseeing the Waymo Driver's operations within a dynamic environment, and to be prepared to assume control of vehicle operation when needed, for instance upon assessment of a situation that may be beyond the current, and ever evolving, capabilities of our system.

Before diving into the specifics of Waymo's recommended FRM framework, it is important to understand the breadth of the challenges that complacency poses to the task of monitoring automated systems. Indeed, clear terminology is needed to assess what the measures presented in this report do and do not cover. Fatigue and drowsiness are interlinked with other driver states, such as distraction, automation complacency and overreliance, all of which contribute to driver inattention. As an illustration of this complexity, to mitigate fatigue and drowsiness, drivers naturally tend to engage in secondary tasks to generate stimulation, potentially leading to distraction-related inattention errors [24]. Thus, inattention related concepts often contain both deliberate and non-deliberate characteristics. Unlike some of the deliberate behaviors that we observed in our early testing (for the older driver assistance technology we had in 2012), this report explores countermeasures that are, for the most part, aimed at addressing risks from *non-deliberate* behaviors, which may ensue when monitoring fully autonomous technology -- the type we currently test and develop.

The United States and European Union Bilateral Intelligent Transportation Systems Technical Task Force (US-EU Bilateral ITS TF), defined a conceptual framework and taxonomy of driver inattention [25] (see Figure 1). Inattention is conceptualized in terms of mismatches between the driver's current resource allocation and that demanded by activities critical for safe driving. Within this taxonomy, inattention is broadly divided into two general categories: (1) insufficient attention and (2) misdirected attention, relating to the activation and selective aspects of attention respectively. For each of these categories, Figure 1 details a set of sub-processes that can give rise to inattention, with fatigue-induced risks affecting the "sleep-related" impairment category.

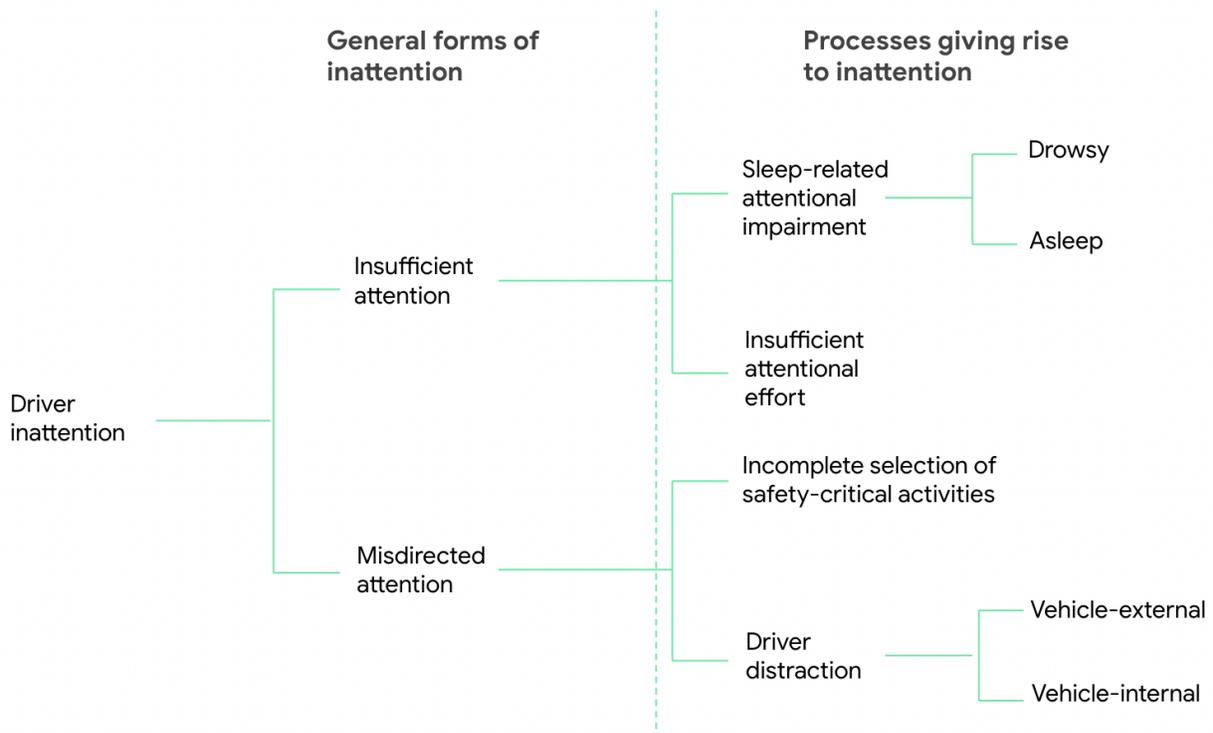

Figure 1. Inattention taxonomy, adapted from [25]

Driver inattention, in all its forms, is recognized as the top contributory cause to U.S. road fatalities. Yet, pinpointing the contribution of inattention to on-road incidents can be challenging. It is often difficult to precisely differentiate between the many inattention-related concepts given the conceptual complexity, measurement difficulties, and overlap in performance impairment; however, they call for different interventions [24]. This makes both estimation of the magnitude of the problem and the evaluation of countermeasures a difficult task. Focusing on fatigue, the 5-year average percentage of fatal crashes and fatalities attributed to drowsy driving between 2011 and 2015 in the U.S. was estimated to be roughly 2.5% [26], but many researchers consider the actual amount to be much higher. In fact, the National Highway Traffic Safety Administration (NHTSA) and the American Automobile Association (AAA) recognize drowsy driving as one of the top three causes of incidents on the road, estimating that fatigue is a contributory factor in over 20% of highway crashes (fatal and non-fatal) [27; 28].

Consensus statements from fatigue research experts state that fatigue contributes to 15 to 20% of transportation crashes [29].

Often, inattention can go unnoticed, even to a trained operator. This is a common factor that has been studied across multiple sectors of the transportation industry, with a special attention within the aviation context [30-32][7]. Recently, one key research development is that of local sleep (see, for instance, [35]). Local sleep, whereby local brain regions can be in sleep mode in awake individuals, may affect brain regions related to motor function such as steering [35]. This illustrates the importance of continuously following research advancements within this complex area to inform and improve the development of countermeasures.

Waymo is well aware of the complexity and ongoing evolution within the scientific field that studies insufficient attention manifestation and countermeasures. At the same time, as a responsible player within the industry, we feel compelled to share what we believe is the state of the art of an FRM program for the AV industry specifically. Our program tackles aspects of insufficient attention, and is in continuous evolution in its implementation, as informed by over 10 years of experience on the road. As stated previously, while the implementation of countermeasures may change (also depending on specific use-cases), we believe the high-level framework here presented has remained stable to the point of providing value to others in this space.

## THE NOTION OF FATIGUE

Even though fatigue is recognized as affecting all modes of transportation [36], it can have particularly dire consequences when it occurs in on-road vehicles. This is due to both the short time intervals available for evasive maneuvers once a road conflict ensues and to the proximity to, and/or vulnerability of, other road users.

A number of studies in the past have investigated driver fatigue and tried to define causes and tell-tale signs that can help recognize its often unnoticed onset [37-40]. Many studies provide variations of the definition of fatigue; however, several common elements for its characterization can be abstracted:

- Fatigue is characterized by a general lack of awareness and degradation in mental and physical performance;

---

[7] Research within the aviation industry has shown that fatigue-induced risks are heightened when passive monitoring tasks are considered. Microsleeps and decreased brain activity have been observed even in highly trained individuals, such as airline pilots, and a substantial body of literature exists that has analyzed these effects since the 1970s (see [33; 34] and references therein).

- Fatigue is subjectively experienced, and its effects can vary from individual to individual;
- Poor sleep hygiene (e.g., lack of, interruptions) is the most cited cause of fatigue;
- Symptoms of fatigue include tiredness, sleepiness, reduced energy and increased effort needed to perform basic tasks;
- From an operational standpoint, fatigue translates into a disinclination to continue effectively performing the task at hand.

Furthermore, the extensive literature on the topic shows consensus that the effects of fatigue do not necessarily depend on the energy expenditure needed for performing the task at hand [37]. This consideration is further developed into the basic distinction between two types of fatigue: *internal fatigue*, and *task-related fatigue*. Figure 2 provides a summary of characteristics for both types of fatigue. Internal fatigue is mostly sleep-related, while task-related fatigue is activity-related [41].

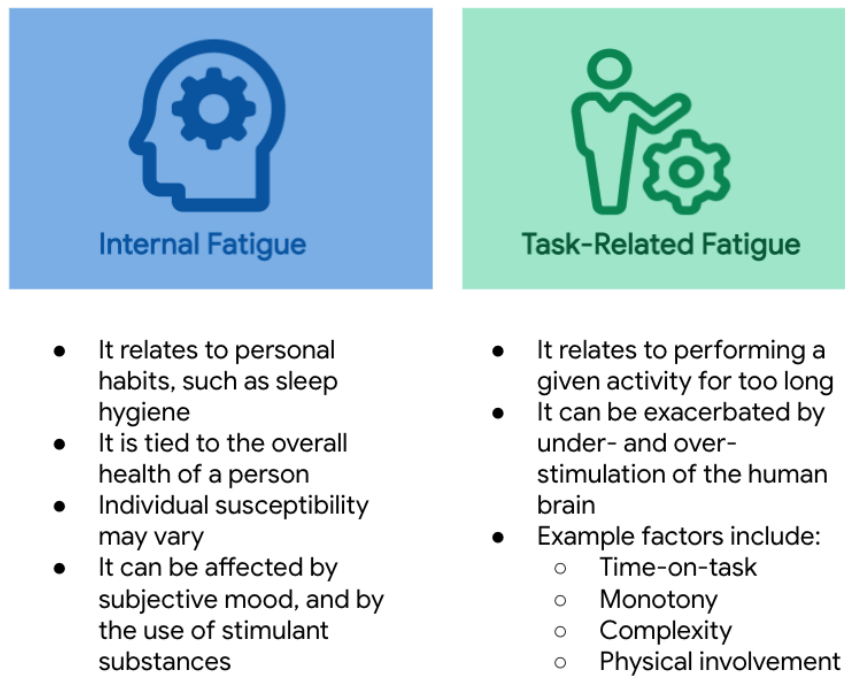

Figure 2. Characteristics of internal fatigue vs. task-induced fatigue

In addition to the characteristics summarized in Figure 2, many additional factors can be conducive to drowsiness, including: the use of certain medications (e.g., antihistamines and benzodiazepines), substance abuse, certain food consumption, and the absence of sufficient exercise [24]. In-vehicle environments can also play a role, for instance in the presence of poor air quality and vibrations [24]. Prior medical conditions may also interfere with sleep (e.g., obstructive sleep apnea), making an individual more prone to internal fatigue.

Both types of fatigue can reduce the alertness of a person, their attention to the surrounding environment, and impair overall performance [42; 43]. Moreover, reduced vigilance can occur well before any feelings or signs of sleepiness are noticed. Literature has shown that internal fatigue and task-related fatigue can occur independently of each other, and can be present at the same time with additive effects [44; 45].

In addition to the distinction between internal fatigue and task-related fatigue, a second layer of differentiation can be added by duration, i.e., how prolonged in time the state of fatigue is. With respect to duration, it is possible to distinguish between *acute fatigue* and *chronic fatigue*.

Acute fatigue, which can be considered a short-term effect, can be a sudden onset of fatigue often due to mental exhaustion. Acute fatigue can range from brief periods to up to a month, and recovery from it may require rest and improved self-care [46].

Chronic fatigue is instead long-term. It is characterized by extreme fatigue that can last for 6 months or more. It is usually tied to periods of intensive stress and to a weakened immune system [47]. Chronic fatigue can further aggravate sleep disorders and post-external malaise, and it is recognized as a debilitating medical condition. Figure 3 summarizes the characteristics highlighted for both types of fatigue.

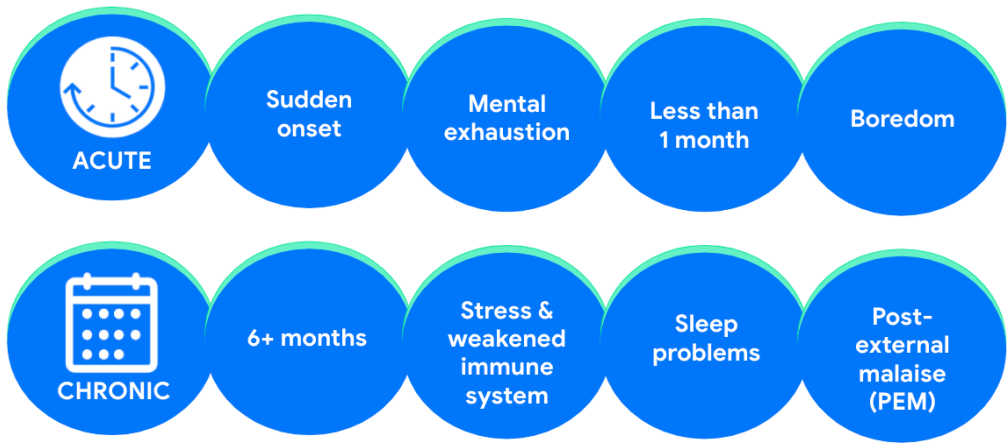

Figure 3. Characteristics of acute vs. chronic fatigue

Regardless of the specific type involved, fatigue can prove disruptive to the well being of autonomous specialists and to the overall safety and efficiency of on-road operations. This is why it's important to understand the operational implications of fatigue, especially in relation to professional driving in the novel automated vehicle space.

## FATIGUE OPERATIONAL IMPACT

The relation between passive monitoring and fatigue is important and relevant to specialists overseeing autonomous technologies. The *"irony of automation"* phenomenon, mentioned before, points to the heightened potential for fatigue for autonomous specialists with an increasingly passive role [48][8]. As noted in Figure 2, the amount of time spent performing a task can be associated with fatigue onset. Decrements in performance have been observed during the execution of time dependent, monotonous supervisory tasks, often leading to a state of hypovigilance [26; 49]. A direct relationship has been observed, where the longer the amount of time spent on unchanging, repetitive tasks, the greater the likelihood of fatigue and of there being a reduced ability to react effectively (see [50] for a review) [24; 26; 51].

Recent research has shown that the evaluation of driver engagement should leverage an additional key component related to cognitive control (i.e., understanding the need for action), rather than more direct indicators of monitoring levels such as purely visual focus (i.e., looking at the threat) or having hands-on-the-wheel [11]. In particular, fatigue-induced risks are known to affect and decrease operators' cognitive control, making their investigation of primary importance at Waymo.

Although Waymo's ADS is L4 and is designed to operate without requiring human intervention, autonomous specialists remain an important part of its development and testing processes, thus making the need to assess how fatigue can affect their safe and efficient operations an integral part of the development process. In particular, the following symptoms related to fatigue may have a negative impact on control takeover in transition from automated driving to manual driving as well as on maneuver executions during manual driving [52]:

- Gaps in attention and vigilance [53; 54];

---

[8] To this point, we would like to draw attention and provide a cautionary warning in the distinction that certain recommended practices bring forth between "early-stage" prototypes and "late-stage" prototypes. Specifically, existing practices indicate that test drivers of late-stage prototypes may not require the same degree of specialized training as those involved in early-stage prototypes [66]. While this statement may bear truth in relation to specific technological limitations of the ADS, its application in relation to fatigue and complacency should be carefully evaluated, given how the improved proficiency of late-stage prototype in executing the DDT may translate in less interactivity on the part of the test driver. In fact, attention to fatigue risks has so far been relegated to short mentions on the need for driver monitoring in testing of prototypes practices both in the US and in the UK [66; 67]. We hope that the present paper will raise awareness of these issues and lead to a more in-depth industry conversation on the impact of fatigue on testing practices.

- Delayed reactions [55];
- Reduced accuracy of movement [56];
- Impaired logical reasoning and decision-making;
- Reduced ability to assess risk or appreciate actions' consequences;
- Reduced situational awareness;
- Low motivation to perform optional activities, which may be tied to assessing fatigue-related status.

Several factors can affect the level of fatigue experienced, and the magnitude of the symptoms listed in this section. Specific factors tied to driving tasks include the ones pictured in Figure 4, which were obtained by combining state-of-the-art knowledge on the effects of fatigue by the National Safety Council, the AAA, and the U.S. Department of Transportation [52; 57; 58].

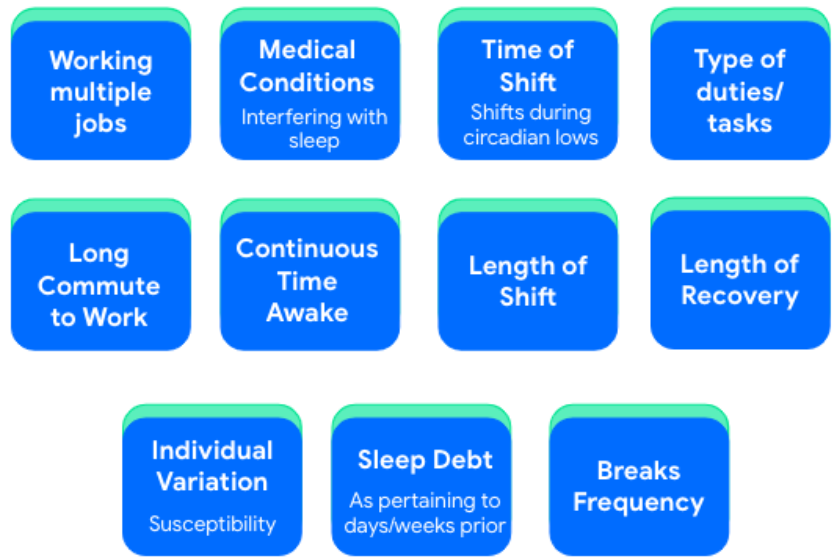

Figure 4. Factors affecting experienced levels of fatigue [52; 58]

Understanding and acting upon factors affecting fatigue is key to the safety of on-road testing operations for autonomous vehicles, and requires the development of a successful fatigue risk management (FRM) program. We review the role of Waymo's proposed FRM framework in the next section, and then dive into the specific blocks and methodologies it encompasses.

## THE ROLE OF FATIGUE RISK MANAGEMENT PROGRAMS

As mentioned in the previous section, fatigue is just one of the recognized sources of vulnerability that humans are subject to while at the wheel. Clear requirements, guidelines, education, training, and surrounding safety culture work hand in hand to counter the effects of inattention. Waymo has a rigorous on-road testing program that has been improved and refined continuously over a decade. Driving on public roads is a critical step that allows us to validate the Waymo Driver's skills, uncover new challenging situations, and develop new capabilities. The safety of on-road testing operations, such as Waymo's, begins with highly-trained operators, who become specialists on safely operating autonomous driving technology, including understanding its current capabilities, and being skilled in strategies to monitor the safety of its operations on public roads. Toward that goal, *autonomous specialists* should undergo extensive education, including classroom training, practice behind the wheel on closed courses, and defensive driving training, before having the privilege to operate and oversee autonomous vehicles. Waymo collaborates with transportation partners who ensure that the autonomous specialists they employ to operate Waymo's autonomous vehicles during on-road testing receive this kind of rigorous training and education, and are consistently fit for duty [24, 52, 67]. Many of these same practices may also help counter distraction, although in this paper we will focus on their relation to countering fatigue-induced risks only.

As new operational design domains are explored, the efforts of autonomous specialists contribute significantly to improve the safe and efficient deployment of autonomous vehicles on public roads and their integration within conventional traffic. During testing of L4 development vehicles, autonomous specialists are responsible for overseeing the system and, if needed, taking control of the vehicle. The FRM framework detailed in this paper thus promotes safe operations, which enables responsible and appropriate development of this technology.

FRM programs are one element of a comprehensive approach to safety, where hazards are identified and appropriately managed. The implementation of an FRM program requires a systematic approach, involving safety requirements and guidelines; incident reporting and analysis systems; and reactive, proactive, and predictive risk assessment. Appropriate objectives for FRM programs need to ensure that risks associated with the identified hazards are managed and mitigated to a level as low as reasonably practicable [59].

The proposed framework is informed by the substantial literature and operational experience within multiple industries: both in the transportation sector (e.g., aviation, railways, and commercial motor vehicles), as well as non-transportation related (e.g., nuclear and chemical) [58; 60-64]. Furthermore, nascent practices within the automated vehicle industry can also be leveraged to benchmark and fine-tune ingredients of an FRM that can prove successful, while remaining practical, during development and testing of an automated driving system. Examples of those include the Automated Vehicle Safety Consortium (AVSC) best practices on in-vehicle fallback test drivers [65], the SAE newly revised standard on on-road testing [66], the BSI publicly available specification for safety operators involved in on-road trialing [67], and a breadth of other literature (see for instance [68-70]). Waymo's FRM framework revolves around three layers, each building deeper defenses against the risks induced by fatigue (see Figure 5):

1. Fatigue Prevention
2. Fatigue Monitoring
3. Fatigue Mitigation

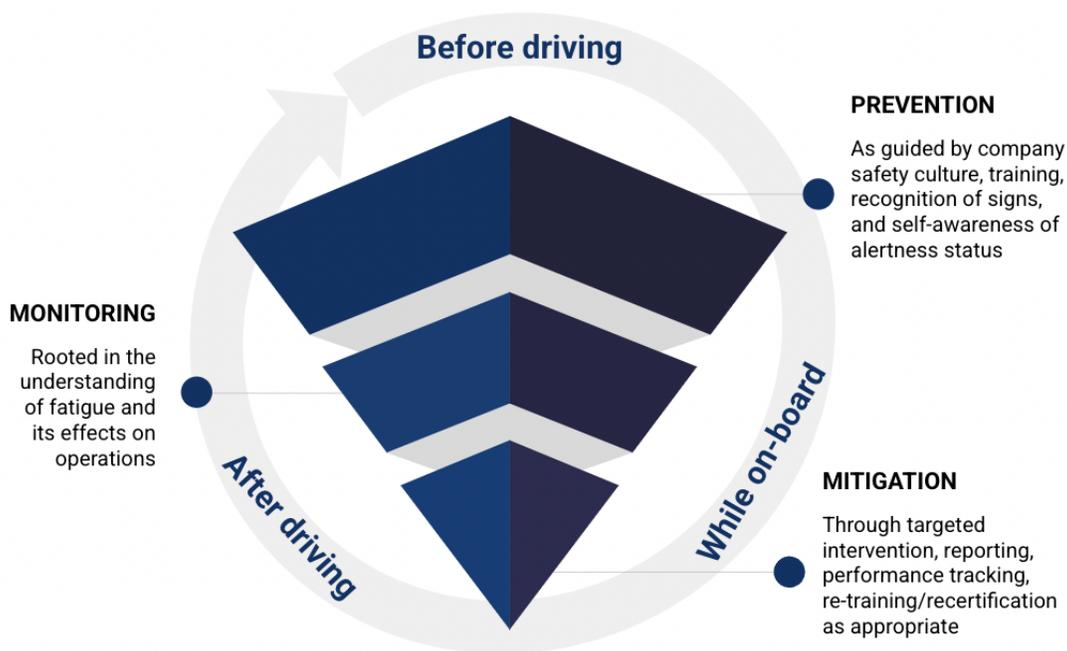

Figure 5. Visual representation of Waymo's layered approach to FRM. Each ingredient should be implemented and assessed at all stages of driving operations

Rather than functioning in isolation, each ingredient encompasses methodologies that feed the FRM cycle at multiple stages, and that span the totality of operations before, while, and after driving[9]. This report explores the ways in which these methodologies can help companies to prevent, monitor and mitigate the possibility of fatigue-related events.

---

[9] Waymo's recommended FRM framework is intended to supplement other safety measures for autonomous specialists and applicable state and federal requirements (e.g., for commercial motor vehicles, hours of service, controlled substance and alcohol testing, and employer screenings).

This framework for FRM applies to both single- and dual-specialist configurations (i.e., testing missions with a single autonomous specialist on board versus those missions with two specialists). As explained later in this report, Waymo's FRM framework recommends that autonomous specialists undergo different training stages (both related to fatigue and to other aspects of autonomous technology testing missions), which allow qualified specialists to operate in a single-specialist configuration for certain software releases. While some may think that the presence of a dual specialist on testing missions could be a consistent and reliable defense against fatigue-induced risks, our experience and observations have indicated that the dual specialist configuration is not a sufficient guard against fatigue without additional fatigue mitigation layers being in place. In fact, early investigations and listening sessions on this aspect identified some key considerations:

- There are challenges in ensuring a reliable monitoring of fatigue markers from the dual specialist. This is because of the sub-optimal position of the dual specialist: to the side and not in front of the primary specialist, with the dual specialist facing forward, unable to have a clear unobstructed view of the driver's face. Furthermore, depending on the specific mission, dual specialists may be preoccupied with other tasks, and may miss subtle or even cumulative fatigue indicators;
- There are challenges associated with peers being tasked with reporting fatigue signs. In fact, significant social pressures can exist between the two specialists, hindering both the accuracy and reliability of peer-based reporting (e.g., feeling social pressure from the co-specialist not to admit to being fatigued or to take an additional voluntary break). Our research and listening sessions identified that specialists felt a certain degree of pride associated with not showing fatigue signs in front of a fellow specialist and a reluctance to report fatigue for colleagues with whom they felt a bond or affiliation with. Furthermore, these sessions highlighted that specialists tended to time their use of additional alertness breaks according to the needs of the fellow specialist rather than based on their own needs;
- There are challenges associated with over-reliance on the opinion of a fellow colleague. We observed that reliance on the judgment of the dual specialist led to a less marked realization of one's own responsibility for self-identifying fatigue. In other words, a diffusion of responsibility took place, where specialists tended to assume everything would be fine until the dual specialist called out signs of fatigue;
- There are challenges associated with ensuring the dual specialist can maintain

their own alertness, especially when assigned the sole task of monitoring the primary specialist.

These considerations led to the design of a framework for FRM that does not rely exclusively on observations from a dual specialist. While dual specialists (when present) should be encouraged to raise fatigue concerns when or if they have them, Waymo's recommended FRM framework views such a role as only one layer in a more comprehensive strategy for mitigating fatigue.

FRM is just one of the many building blocks that inform and make up the operations layer of Waymo's safety methodologies — the third layer presented in [22]. In particular, FRM is part of the safety practices surrounding the fleet operations block presented in [22]. As such, it ties with a host of other methodologies and practices, not covered in this report, related to ensuring that autonomous specialists achieve:

- the appropriate understanding of the performance capabilities of automated driving systems;
- the appropriate understanding of the specific role that specialists play in targeted aspects of the development of this technology;
- the appropriate understanding of additional challenges (other than fatigue) associated with remaining attentive to their task;
- the appropriate understanding of incident management procedures.

The remainder of this report is organized as follows. Section 2 presents Waymo's high-level framework for FRM and details of the various methodologies that make it up. Given the breadth of these methodologies, the reader is first introduced to our philosophy and the overall organization of the constitutive blocks, then presented the details of the methods, and then finally brought once more to the high-level overview that frames the FRM as a whole. Section 3 concludes this report with a short discussion on lessons learned, and additional considerations tied to the aspirational goals behind sharing our FRM framework with the public.

# 2. Waymo's Proposed Approach to Fatigue Risk Management

## FRAMEWORK PRESENTATION

Waymo is committed to reducing traffic injuries and fatalities by driving safely and responsibly and to carefully manage risk as we scale our operations. As part of Waymo's pledge to prioritize safety, Waymo welcomes feedback and suggestions for improvement from all those involved in our operations, including autonomous specialists. Speak-up attitudes are empowered by clear safety concerns-reporting mechanisms. In kind, providing education on fatigue and Waymo's technology and processes for mitigating in-vehicle fatigue risk, remains a foundational part of Waymo's core support for safe and efficient operations. These recommended practices help foster a culture of good habits and self-care, and minimize exposure to the risks induced by fatigue.

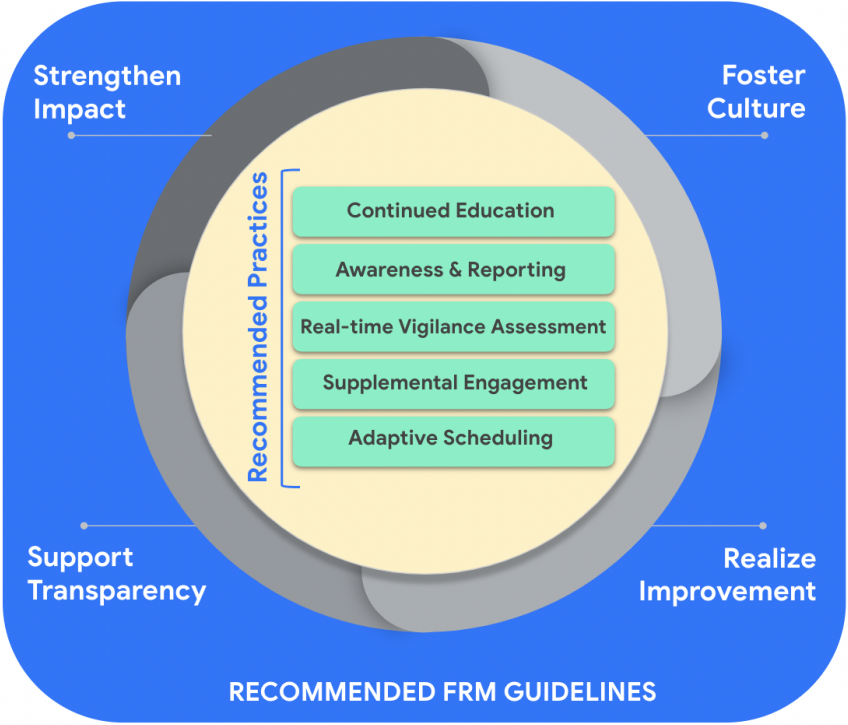

Figure 6. Notional representation of Waymo's Fatigue Risk Management framework

The careful analysis of the background presented in the previous section led Waymo to identify a number of methodologies and elements for this proposed approach to FRM. The resulting framework, informed by years of experience of testing fully autonomous vehicles on real-world roads, is notionally represented in Figure 6.

Before presenting the details of the methodologies that constitute this FRM framework, it is necessary to understand how those come together in a unified and cohesive program. Waymo has distilled four underlying goals that touch upon all the aspects of a FRM program (represented by the outer circle of Figure 6), and that serve as both the north star of our efforts and a practical guide:

1. Foster Culture: Waymo is committed to building and promoting a strong safety culture within and beyond our organization, to reach all our customers and those that share the road with us.
2. Realize Improvement: A safety culture can only be said to exist when the value of improving safety education and performance is widely shared across all employees and transportation partners.
3. Support Transparency: Achieving improvement starts with an honest and open discussion of where we can improve and do better, and progresses with a humble approach to continued learning.
4. Strengthen Impact: We strive to create structured processes to ensure that the implemented methodologies achieve their intended objectives, and to improve them where they fall short. We continuously learn from our experience, and inform changes to our practices, when appropriate, to achieve our goals.

These four goals feed into the methodologies that embody Waymo's FRM framework and the associated establishment of clear requirements and guidelines - the basis and foundation on which the framework stands. These methodologies, presented in the next section, tailor a number of industry-recognized fatigue risk management best practices, and introduce novel patented technological techniques. The methodologies that comprise Waymo's FRM framework are organized under five main headings (or "implementation blocks"), presented in the next section:

- Continued Education: inclusive of education on autonomous driving technology and FRM practices;
- Awareness and Reporting: inclusive of voluntary and mandatory practices for fatigue reporting, awareness, and self-assessment;
- Real-Time Vigilance Assessment: inclusive of both automated and human-based monitoring procedures, grounded within Waymo's drowsiness rating model;
- Supplemental Engagement: inclusive of notifications and in-vehicle operations to facilitate autonomous specialist engagement while monitoring the operation and performance of autonomous technology;
- Adaptive scheduling: providing resources to help autonomous specialists' employers decide effective shifts rotations and resting periods, flexible breaks, and strategic tasks reallocation to minimize fatigue events.

# IMPLEMENTATION BLOCKS

*Continued Education*

The first block of the FRM framework involves providing resources on continuing training and education on fatigue management for autonomous specialists (AS). This continued education block is inclusive of bi-directional learning processes (Figure 7):

- Training and education of autonomous specialists (to AS)
- Feedback and lessons learned (from AS)

Specialists should be provided with both initial and ongoing guidance for fatigue prevention, monitoring, and mitigation, to promote alertness and awareness of fatigue-related risks at all times. Starting from fatigue prevention best practices (including self-care and good habits strategies), the approach continues with education on monitoring signs of fatigue, combating them, and mitigating fatigue-induced risks (which includes becoming familiar with the overarching FRM program and all the recommended implementation blocks described in this section).

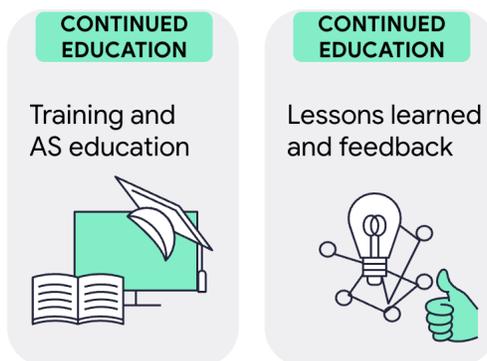

Figure 7. Elements of the continued education block

Educational programs should happen on various timescales, and extend to all those that deal with vehicle operations: from autonomous specialists and trainers, to support personnel, testing and calibration engineers, and all who participate in the monitoring of vehicle operations. While those dealing with vehicle operations are the focus of these programs, Waymo recommends making the core fatigue education material available to all employees, as well.

It's recommended that autonomous specialists are exposed during the initial stages of training to specific fatigue content, including: the notion of fatigue, its causes and markers, fatigue mitigation techniques, and fatigue monitoring and self-awareness.

Where necessary, more frequent fatigue training should be provided, as determined by the specialists' employer. This line of training is an additional layer of education and skills-refinement on top of a more traditional defensive driving training and ADS-specific training that each specialist must undergo before being placed at the wheel of an autonomous driving vehicle.

As mentioned in the Introduction, autonomous specialists are an important part of autonomous driving technology testing programs. After the first initial stage of training is successfully concluded, it is recommended that new specialists be first introduced to operations with a secondary (or dual) specialist, before being assessed and given gateway assignments toward single-specialist assignments. In addition, specialists who exhibit frequent and/or severe fatigue events should be re-assessed by their employer and provided additional support as needed before transitioning to single-specialist assignments.

Retraining and recertification present additional countermeasures for addressing fatigue events, depending on frequency and magnitude at which fatigue is experienced. Experiencing fatigue, while a normal and expected part of life, should not happen during vehicle operations, and can be an indication that preventative strategies were not carried out before the commencement of a driving shift. Awareness and self-assessment are also important steps within a FRM program, and specialists should be educated on these practices starting from the day they are first screened.

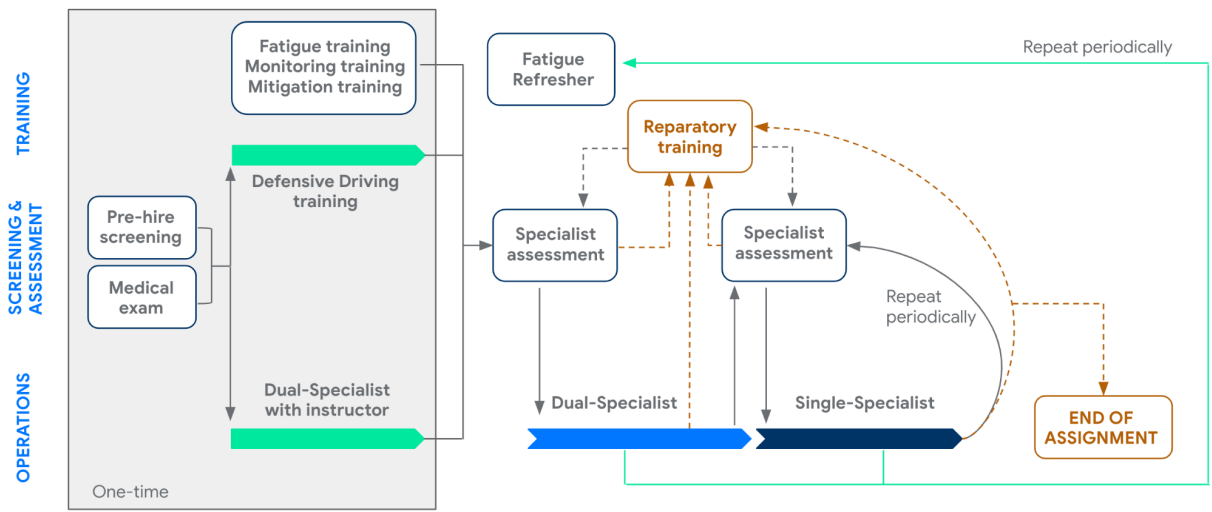

Figure 8. An example overview of training and operational counterpart schedules

Figure 8 presents a diagram that summarizes possible training and operational counterpart schedules for autonomous specialists under a recommended FRM framework.

Continued education practices do not impact autonomous specialists only. In fact, continued education can provide an intrinsic value to an FRM program and safety culture. As part of continued education practices, feedback and flagging of issues within the FRM program should be welcomed. The use of listening sessions, surveys, and open discussion forums, in collaboration with autonomous specialists and their employers, can all help foster a transparent speak-up culture (for instance, to ensure the smooth roll-out of new operational changes).

Feedback from autonomous specialists on the perceived effectiveness and usability of fatigue risk prevention, monitoring, and mitigation techniques can improve the development and implementation of FRM programs. For example, the results from anonymous, voluntary surveys (see "awareness and reporting" element below) can be used to explore causes of previously reported fatigue. Lessons learned can also be shared in company-wide settings, through internal blogs and newsletters. These practices are broadly aimed at tackling the risks induced by fatigue in an open, honest and serious way.

*Awareness and Reporting*

Waymo's FRM framework builds on pillars of education and transparency. Success of these efforts is tied to the empowerment of autonomous specialists to build knowledge and skills to assess when they are experiencing fatigue, and to the freedom to candidly report when they are experiencing fatigue. The awareness and reporting block includes two recommended methodologies (Figure 9):

- Periodic Fatigue Surveys (PFS)
- Safety concern escalation paths

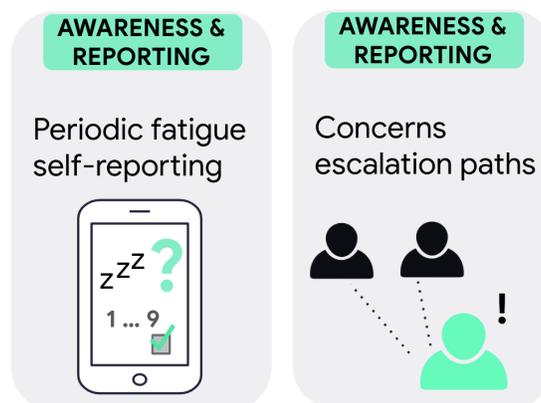

Figure 9. Awareness and Reporting elements: PFSs and Concern escalation paths

A PFS is intended to enable autonomous specialists to self-assess their alertness level when they begin or resume overseeing the operation of autonomous driving technology.

A PFS aims to accomplish several objectives:

- Provide autonomous specialists with a tool that helps them become increasingly aware of their susceptibility to fatigue, as well as improving the skill of self-assessing their fatigue level;
- Create a simple and accessible way for autonomous specialists to report fatigue to their supervisors when or if fatigue occurs;
- Reveal fatigue self-reporting trends to continuously evolve training resources and inform the development of novel and/or better educational approaches;
- Provide a large-scale, longitudinal dataset to help discover which methods are best at promoting the self-reporting of fatigue.

A PFS, which on a practical level takes the form of an online survey that specialists can complete on an individual device, is a simple and accessible method for reporting fatigue to their supervisors and getting assistance in both the short and long term. It is not, however, intended to be used by the specialists' employer as a performance metric. More specifically, within each PFS survey, a specialist reports a subjective, self-assessment of their fatigue or alertness level.

The subjective assessment of alertness follows the Karolinska Sleepiness Scale - KSS: a 9-point Likert-type scale, ranging from extremely alert to extremely sleepy [71; 72]. The assessment refers to the level of sleepiness experienced in the 5 minutes prior to the commencement of the survey. The reporting of scale level of 6 (i.e., "some signs of sleepiness") or beyond is an indicator to the autonomous specialist and their supervisor that the specialist should consider taking an additional alertness break[10], after which the specialist should be prompted to submit a follow-up survey. In cases where alertness has not improved following the alertness break, the supervisor should have the discretion to reach out to support the autonomous specialist as appropriate. Tips for autonomous specialists may also be suggested at this time, which may include physical and/or mental activity aimed at improving the specialist's level of alertness. Figure 10 provides a sample trendline over a specialist's shift. Analysis of PFS data, in aggregate fashion, may help specialists' supervisors identify patterns of fatigue self-reporting allowing them to compare different routines, shift schedules, and driving configurations. Furthermore, honest and frequent self-reporting can aid the development of fatigue

---

[10] It is recommended that autonomous specialists may also elect to take an alertness break for reported values lower than 6, as well.

countermeasures.

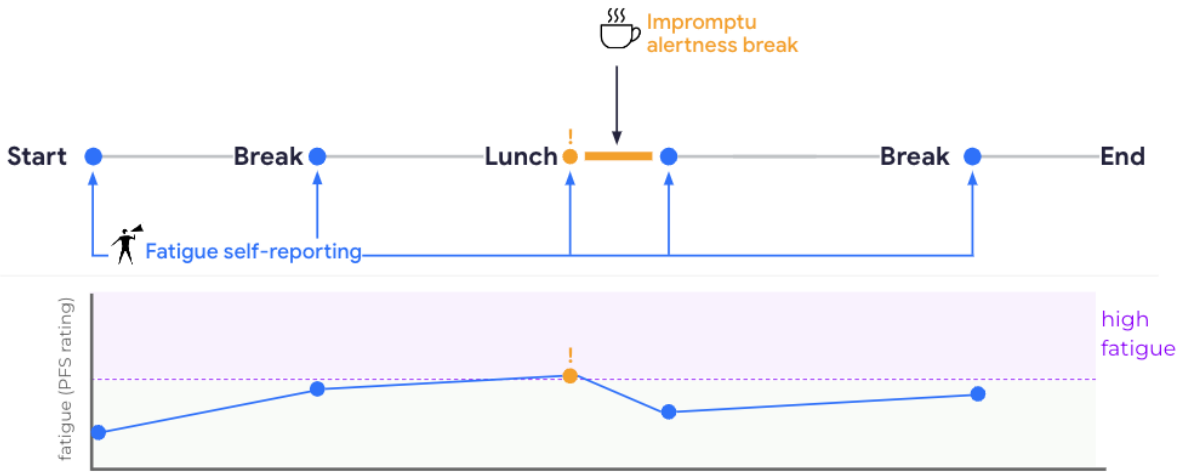

Figure 10. Notional example and representation of PFS trends. For illustrative purposes only.

As explained before, a PFS is a subjective assessment and, as such, it balances two well-known aspects: on one side, given that fatigue signs are highly variable across individuals, it provides an important first-hand account of subjective experience; on the other hand, due to its subjectivity, it may be a less reliable/consistent tool. This limitation may be addressed in a number of ways:

- By striving to facilitate consistent and honest reporting from specialists through positive reinforcement by their employers and guidelines for PFS self-reporting;
- By clearly conveying that the level of fatigue reported in a PFS is not used to assign a formal drowsiness rating to the specialist, preventing an incentive for reporting bias. Formal drowsiness ratings may instead be computed through both human and automated fatigue monitoring, done while in the vehicle, and associated with a separate metric for the specialist (as explained later within the "vigilance assessment" section). A PFS used to quantify performance may incentivize over-reporting of alertness, which would undermine the utility of this method;
- By providing consistent training on the use of PFS through a set of "read/observe-comprehend-assess" canonical examples of fatigue accounts, ensuring a thorough comprehension of this fatigue scale prior to real-world use.

Self-reported fatigue data can be used to evaluate the efficacy of other fatigue measures too. For example, the fatigue classification of an automated fatigue monitoring system may be compared to a subsequent and proximal PFS self-reported instance of fatigue to assess whether an automated system accurately detected the self-reported instance of fatigue.

When the results from PFSs provide an indication of high-fatigue, or upon additional circumstances raised by the autonomous specialist to their supervisor, the specialist's supervisor may have the discretion to consider redirecting the specialist to a different task (see the "auxiliary task assignment" element within the adaptive scheduling session) or take proactive steps to work with their specialist to ensure adequate rest and recovery before their next shift.

A PFS is only part of the recommended solution for evaluating specialists' alertness levels. The importance of self-care and good personal habits should be stressed from day one. While training may be a key part of ensuring good habits are set from early on, a robust safety culture may also help to encourage specialists to carry these practices into their own lives. This is why the second pillar for the awareness and reporting blocking is especially important. The use of multiple diverse paths for escalation of concerns helps ensure autonomous specialists are provided with readily available and easy to access reporting mechanisms. In fact, autonomous specialists may vary in their comfort and willingness to escalate safety concerns through specific channels (e.g., through a supervisor).

For the example case of a PFS, such escalation may occur as more of a direct and open communication between autonomous specialists and their supervisors, which serves as an additional layer of fatigue mitigation and can offer opportunities to raise specific fatigue circumstances before and during a shift. Indirect pathways exist too. Additional programs like the usage of anonymous voluntary surveys, as mentioned in the continued education block, provide another means of escalation. Concern escalation paths are, however, not just a way to elicit feedback from autonomous specialists. Other personnel involved with or with exposure to operations may feel the need to escalate safety concerns, so that more general concerns escalation paths can exist to solicit such input. For example, Waymo's Field Safety Program helps collect, assess and resolve potential safety concerns that occur during real-world operations from many sources, including employees, transportation partners, our riders, and the public [22].

***Real-Time Vigilance Assessment - Integration of Automated Fatigue Monitoring with Human Remote Fatigue Monitoring***

While many of the recommended methodologies related to screening, training, and educational revisions are tied to prevention and/or mitigation of fatigue-induced risks, it's also important to understand how to identify, rate the level of, and act upon, fatigue in real-time. As explained below, when situations of high fatigue are observed, these recommended FRM practices also account for appropriate steps for intervention.

Waymo's FRM framework recommends assessing the level of drowsiness or fatigue experienced by a given specialist through two concurrent processes (Figure 11):

- An automated monitoring system
- A human-operated monitoring system

An automated fatigue monitoring system (also often referred to as an eye tracker, or camera-based driver monitoring system (DMS)) typically faces the driver and uses eye closure, head movement, and eye movement to detect fatigue and distraction. This type of monitoring is sometimes called "direct" DMS (as opposed to "indirect" DMS, which infers driver fatigue or distraction from, e.g., lane keeping behavior or hands on wheel detection). Upon real-time detection of fatigue or distraction, an automated fatigue monitoring system may issue multi-modal alerts to the autonomous specialist, consisting of, for example, a tone, a vibration under their seat, and a flashing light. It may also then provide a real-time video feed to remote human raters of fatigue.

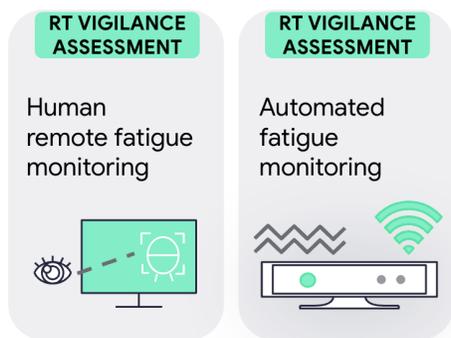

Figure 11. Real-Time (RT) Vigilance Assessment elements: drowsiness ratings from human and automated monitoring

Both human and automated ratings rely on a clear definition of a "driver alertness model". Such a model characterizes the distinctive features associated with fatigue, which are then used for both automated and human ratings to assess the drowsiness of autonomous specialists. The driver alertness model is thus at the base of the monitoring techniques recommended as part of this FRM framework. Waymo has used models inspired by established literature (see [40; 73; 74] and references therein), and the "observer rating of drowsiness" (ORD) protocol originally developed by the Virginia Tech Transportation Institute (VTTI). Additionally, advice provided by VTTI experts helped inform subsequent efforts to refine aspects of this framework.

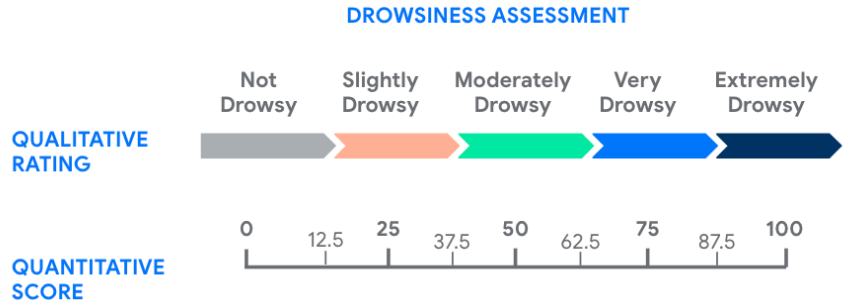

Figure 12. An example of a drowsiness assessment scale, adapted from [40; 74]

A driver alertness model may consist of five qualitative and quantitative levels of drowsiness. Examples of observer-based qualitative ratings and quantitative ratings are shown in Figure 12, which provide an overall assessment for the level of drowsiness experienced by an autonomous specialist at a given point in time.

Table 1 provides examples of summarized factors and indicators subsumed under each of the five drowsiness levels. The descriptive mannerisms associated with each category are central to the rating provided in both the automated and the human-operated process. Trained human raters examine the presence/absence of these drowsiness-related mannerisms. Although assessments of driver distraction may differ from assessments of drowsiness in approach and goals, it may be helpful for drowsiness raters to also flag potential sources of distraction, such as identifying the presence/absence of unsafe secondary tasks (e.g., personal phone use).

Table 1. Examples of progressive drowsiness indicators and mannerisms, adapted from [73]

| Not Drowsy | Slightly Drowsy | Moderately Drowsy | Very Drowsy | Extremely Drowsy |
| --- | --- | --- | --- | --- |
| Exhibition of alert behaviors, such as normal fast eye blinking, short glances to surroundings. Occasional body movement and gestures may be observed | Looks may be less alert and sharp. Glances can become longer and eye blinks may not be as fast | Start of exhibition of efforts to combat drowsiness, such as face rubbing, eyes rubbing, scratching, facial contortions, restless movements. Some individuals may appear more subdued, and may stare at fixed points | Clear signs of drowsiness may be observed, such as prolonged closures of eyelids (2 s or longer), rolling upward/sideways movement of eyes. Some individuals may display a lack of apparent activity | Specialists may be falling asleep. Prolonged closures (over 4s) of eyelids may be observed. Similarly, periods of inactivity are observed. Large body movements may be observed in conjunction to transitions in/out of dozing |

The indicators of Table 1 can be more formally captured within the user interface employed for drowsiness ratings as part of a human-operated system. An example of a graphical interface for rating is presented in Figure 13, which may feature a Likert-type scale, rated following the identification of fatigue indicators in the following categories:

- Eyes
- Face/Head

- Body

For each category, several of the indicators from Table 1 are listed for identification. Additionally, the rating interface may feature a separate observations category which may contain things like mobile device use[11] and hands' placement. While specialists' engagement with secondary tasks are not used as indicators of alertness/fatigue per se, the secondary engagement may have an impact on other indicators of drowsiness (e.g., slow movement in changing a radio station).

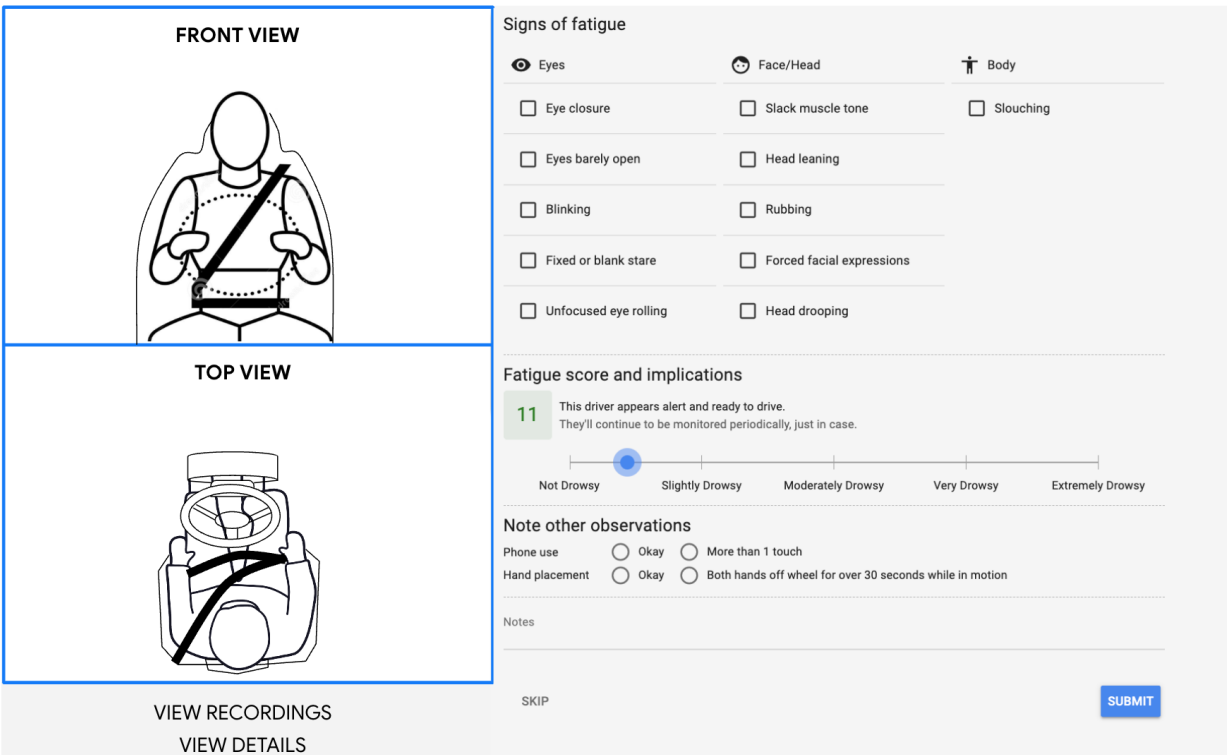

Figure 13. An example user interface utilized in a human-operated drowsiness rating system

As indicated in Figure 11, a system like this may employ a combination of automated and human-executed live monitoring, with human raters' observations as the final fatigue assessment. Depending on the source of the rating, two routes are possible:

- Within route one, when the automated monitoring system has immediately alerted the specialist, it also escalates for observation the video feed of the potentially fatigued specialist to multiple human raters, who each independently rate fatigue in order to validate the automated flag.

---

[11] Personal devices should be prohibited from use while operating autonomous driving vehicles. To the extent work devices are used in autonomous driving operations, they should have restrictions in place that limit functionality and enhance usability with limited specialist interaction in appropriate situations.

- Within route two, when a single remote human rater assigns a high fatigue rating through periodic observation, immediate action should be taken by a supervisor to address the potential fatigue event and a multi-rater validation of that observation should be executed by additional human raters.

Across both routes, the multi-rater validated rating should be considered more accurate than a single-rater estimate of observed drowsiness[12]. An important feature of this process is that the human raters should be independent and unable to distinguish between video feeds that have been escalated for cross-validation and those that are a part of the regular, periodic monitoring of all drivers; this helps prevent a bias that could occur if raters knew the driver had already been identified as fatigued by either the automated system or another rater. Cross-validation with human raters also helps assess and improve the accuracy of the automated monitoring processes, where automation helps achieve continuous monitoring of vehicles' specialists and serves as an added layer to mitigate fatigue-induced risks. As more data are collected, the automated assessment algorithms and the specialist alertness models can be fine-tuned and refined. Previous research has highlighted challenges associated with observer rated sleepiness methods [75; 76], which is one of the reasons why this recommended fatigue monitoring approach is multi-pronged and leverages a combination of multiple and often redundant metrics. Note that the use of multiple raters provides the option to measure reliability of fatigue ratings: reliability of fatigue ratings can be evaluated across raters at regular intervals, and the quantity of raters needed for sufficient cross-validation can be adjusted accordingly. A similar process can also help identify raters that may benefit from additional training. Furthermore, a system that employs high resolution video, with consistent IR light sources across day and night may improve the reliability of these ratings [75].

Using this approach, or one like it, trained human raters monitor a real-time (or near real-time) stream from two camera feeds of the autonomous specialist before providing a rating[13]. Raters have the opportunity to escalate, on the spot, technical issues that prevent them from rating a specialist. When situations of high fatigue are witnessed, they would be escalated to a dedicated point of contact, such as a supervisor who would check in with the specialist to provide support. The supervisor may then recommend interventions, such as an additional, alertness break or, in more serious cases, they or the specialist may request that the specialist and their vehicle be retrieved. Interventions may also come in the form of suggested activities (e.g., light to moderate physical activity during an alertness break, or social interactions involving conversation, singing, or a type of play), or adaptations of the operational environment, such as changing music. With this approach, autonomous specialists would have continuous access to support functions (e.g., supervisors), and would be encouraged to reach out when or if fatigue occurs. If fatigue became recurrent, the specialist's employer could, at their discretion, consider conducting an internal review of the specialist's scheduling (see the "adaptive scheduling" block) or providing supplemental training to ensure the specialist is aware of the methods and benefits of good self-care, such as having good sleep hygiene. If these supportive actions were not effective, the specialist's privileges to operate autonomous driving vehicles could then be modified or possibly suspended, depending on the circumstances of each individual case.

---

[12] At any point in time, an autonomous specialist flag of high-fatigue should take precedence to allow for immediate safety interventions, as discussed in this report.

[13] It is recommended that human raters undergo an assessment that ensures both comprehension of the fatigue rating methodology and the consistency of their ratings with a highly vetted test set, which includes real world videos of people experiencing varying levels of fatigue.

*Supplemental Engagement*

The framework blocks discussed so far provide thorough prevention and monitoring countermeasures for fatigue-induced risks. However, they are not sufficient on their own. Data analysis of over 10 years of on-road operations helped us realize that active intervention countermeasures should also be part of a comprehensive FRM. Supplemental engagement is thus aimed toward in-vehicle physical and cognitive engagement of autonomous specialists. This block includes two main methodologies (Figure 14):

- Interactive Cognitive Tasks, or In-Car Tasks (ICT)
- Secondary Alert (SA) of manual driving

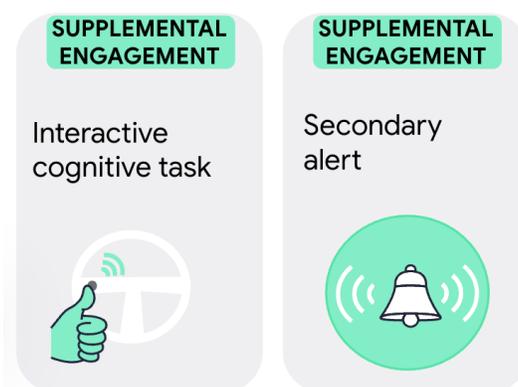

Figure 14. Elements within the supplemental engagement block: ICT, and SA

*Interactive Cognitive Tasks*, or in-car tasks, (ICTs) are tasks designed to prevent and/or reduce the effect of fatigue related to passively monitoring autonomous vehicles.

Research has suggested that across a wide variety of tasks, optimal human task performance (in this case, driving) is achieved with moderate levels of stimulation and task workload [24], as defined by the classical inverted U-shaped function between arousal and task performance [77].

Under this approach, autonomous specialists would experience two types of in-car activities: specialist-initiated tasks and dynamically-requested ICTs. Specialist-initiated tasks are completed at their own discretion, and these may include unprompted commenting, labeling, and evaluation of the driving and contexts the specialists observes[14]. Dynamically-requested ICTs, hereafter simply referred to as "ICTs", are voluntary auditory prompts, introduced algorithmically, only at times when specialist-initiated tasks, and interactions of any sort, have become infrequent. ICTs have been found to improve driver performance and mental state [78; 79]. However, prior research on ICTs has demonstrated that the alertness benefits of ICTs are short lived (e.g., [78]); therefore, algorithmically triggered ICTs are used to provide periodic, supplemental engagement at a rate and level proportional to monotony. It should be noted that specialists must have the authority, capability, and responsibility to disregard any ICT, of any type, at any time, particularly where needed to ensure safe operation of an autonomous driving vehicle. ICTs are a dynamic approach to both prevention and mitigation of fatigue, and are designed to engage both a specialist's motor skills and cognitive skills in their response. ICTs may consist of engaging with available human machine interfaces (HMI), for example by pressing a specific button on the steering wheel in response to a question about the present driving conditions.

ICT requests to specialists can be triggered by different mechanisms and may depend on the specific type of ICT chosen. For example, mechanisms could include:

- A gap in interactivity between the autonomous specialist and the autonomous driving vehicle. Such a gap, which can be measured in terms of both time and/or traveled distance, accounts for the absence of inputs to/from the specialist (e.g., comments on the status of the vehicle, or notifications prompted when a customer hails the autonomous system)[15].
- A failure to execute one or more prior ICTs, either due to incorrect execution or to excessive time taken to complete the task (only applicable to some types of ICTs).

---

[14] Commenting may consist of verbal, open-ended feedback on driving. Labeling, by contrast, is generally mission-specific and may consist of identifying objects and/or observed behaviors. Evaluations apply to driving quality along a dimension of interest.
[15] Note that variance may be intentionally introduced in this process so that prompts become less predictable to seasoned autonomous specialists.

Failure to properly execute one or more ICTs may trigger a follow-up ICT, as mentioned before, while a second miss may trigger an intervention. Mitigation techniques may include one or more of the following: direct communication with support personnel; starting video monitoring streams to assess the situation; and HMI activation (e.g., an auditory alert). Support personnel may also suggest countermeasures that promote a specialist's alertness such as, for example, taking an alertness break. If an ICT's performance indicates that fatigue may be high, more urgent countermeasures, such as pulling the vehicle over, may be initiated.

Additional triggering mechanisms than those described above are also being explored, and a team of Waymo engineers, human factors and psychology researchers are pioneering technological advancements tied to ICTs, and continue to investigate various ways to implement those advances in the future[16].

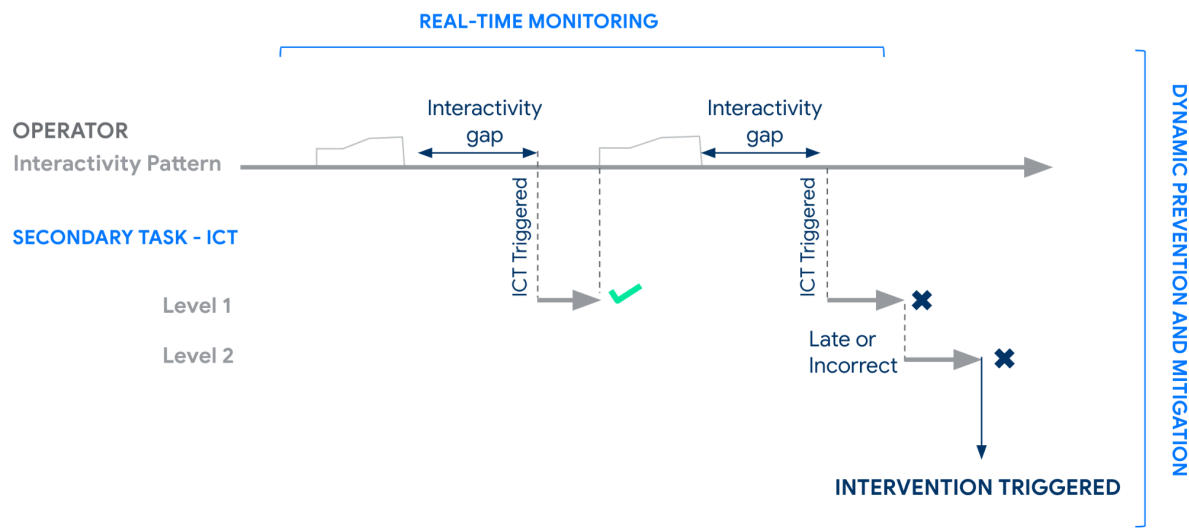

Figure 15. Visual schematic of example ICT functioning, applicable to simple ICT types that autonomous specialists would be expected to satisfactorily complete under most circumstances

In general, both the frequency and complexity of ICTs may be dialed up or down, depending on the level of involvement of a specialist in the primary task (i.e., overseeing vehicle behavior and/or driving, depending on the mode of operation of the vehicle) at the point in time of interest. Furthermore, the frequency of ICTs and their triggering criteria can be customized for each autonomous specialist, based on their individual history, such that specialists with recent errors/misses or slower response may be subject to more frequent ICTs.

---

[16] *See, e.g.,* U.S. Patent No. 10,807,605; U.S. Patent Publication 20210001864; U.S. Patent Publication 20210001865; WO 2020/131803; U.S. Patent Publication 20210106266; WO 2021/072064; U.S. Patent Publication 20210046946; U.S. Patent App. No. 16/928,630; U.S. Patent App. No 17/167,289.

Figure 15 provides a representation of a possible ICT implementation, spanning prevention, monitoring, and mitigation (i.e., intervention in this case) phases.

*Secondary Alerts* (SA) of manual driving are, as the name implies, a secondary or follow-up alert of the vehicle control transition from automated to manual, which follows a primary alert triggered at the time of control transition. A secondary alert is employed to reduce the risk of an unintentional transition from automated to manual (e.g., accidentally tapping the accelerator after hitting a pothole) going unnoticed (e.g., due to fatigue or distraction).

Under this approach[17], an autonomous specialist overseeing the automated driving system operations has the ability to revert from automated to manual control in a number of ways, ranging from pressing a dedicated button on the steering wheel to providing an appropriately calibrated input in the form of steering/braking/throttle. When the control transition occurs, it triggers a primary alert to notify specialists of the change in control authority and the need to resume manual control. Yet, inadvertent transitions of control may happen, so secondary alerts can provide an added layer of mitigations in situations where an operator may be subject to fatigue. Similar to ICTs, SAs may be triggered based on timing considerations tied to a period of inactivity (both before and after the primary alert), as explained next. Once an SA is triggered, it may be cleared by appropriate input by the specialist; if the specialist fails to provide such input, an alert may be sent to support personnel.

The SA logic is based on the likelihood of the request for control transition being unintentional or unnoticed. It can consider a variety of factors, such as the presence or absence of specialist input before and after the control transition, the vehicle's state (e.g., speed), and how the specialist took control (e.g., button vs pedal). The use of a variety of factors allows the SA logic to be tailored based on patterns of risk and expected cognitive load. This allows, for instance, to make the determination of whether an SA may be unnecessary in an emergency situation, where it may be deemed distracting for an effective control take-over (i.e., contributing to cognitive overload).

---

[17] During testing, the disengagement logic may balance gaining confidence in the intention to disengage the vehicle (i.e., certainty that a disengagement was intended by the specialist) against the latency/delay before allowing the control transition to happen (i.e., how fast the specialist can resume control when they request it). While simple solutions such as a constant delay or threshold may come to mind, those may not work in other safety relevant scenarios that may be encountered during testing stages.

*Adaptive Scheduling*

Adaptive scheduling practices are another tool that employers can utilize to further mitigate the fatigue experienced by their autonomous specialists. Adaptive scheduling elements include recommended practices that target many of the factors that Figure 4 linked to the level of fatigue experienced by a specialist. Elements within this block include (Figure 16):

- Shift-forward rotations
- Smart break scheduling
- Auxiliary tasks assignments

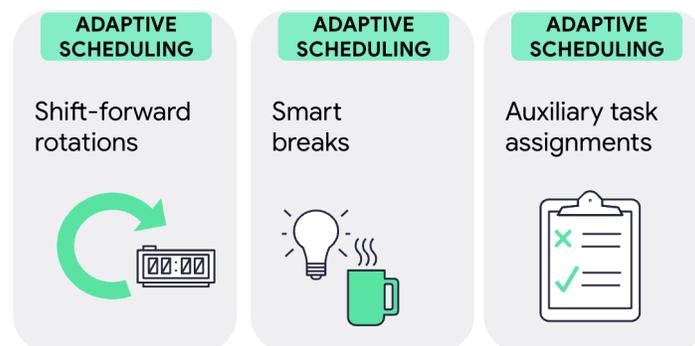

Figure 16. Adaptive scheduling elements

Sleep/rest scheduling practices may also be part of recurrent training for autonomous specialists. Stable routines are recommended, but in situations where specialists need to, across successive shifts, transition the shift start and end times, employers may want to transition shifts in a forward and gradual fashion, which can benefit sleep quality relative to shift-backward rotations [80].

For example, in *shift-forward rotations*, work periods may be incrementally later across each successive day, or may be separated by extended time off when transitioning to a new earlier shift. An example is provided in Figure 17, where the shift start-time is slated to be 4 hours later, increasing over two days, moving start-times forward two hours each day. In contrast to the recommended shift-forward approach, in backward schedule rotations, start-times are instead moved earlier in the day, often resulting in challenging changes in routines and fatigue. Additional resting periods may be needed to adjust to new shifts or schedule changes.

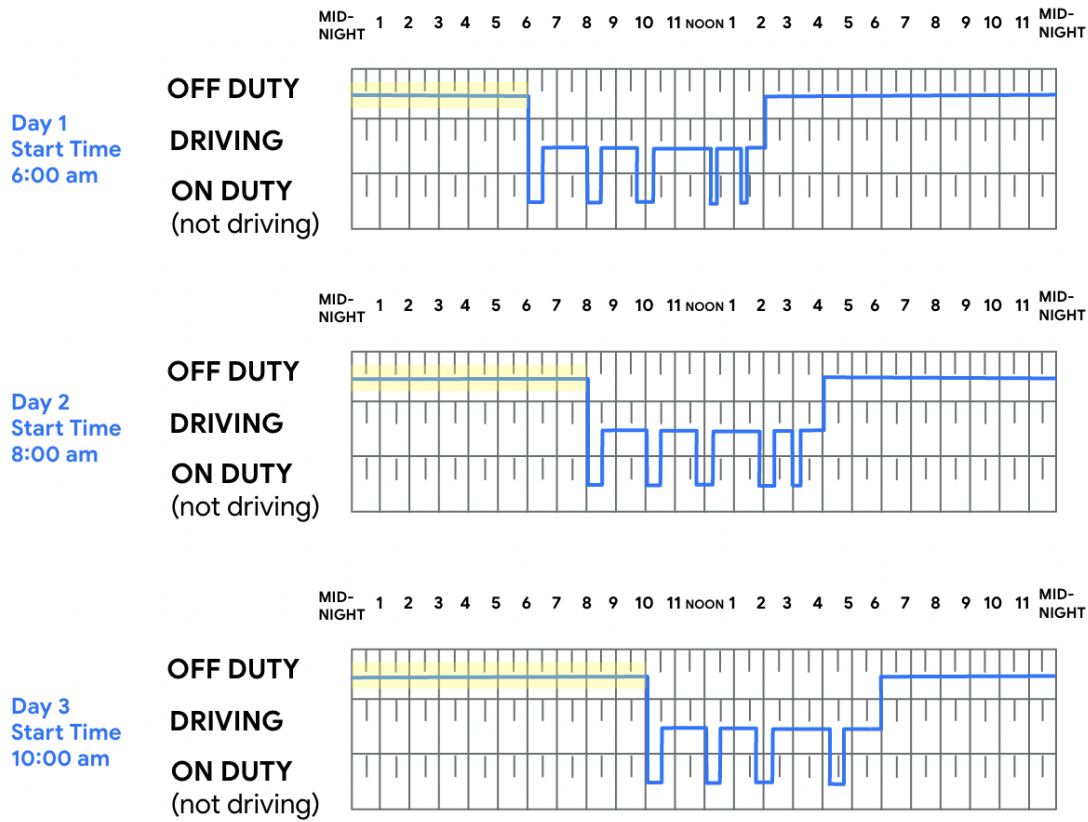

Figure 17. Sample schedules of three consecutive days with forward rotation. For illustrative purposes only.

The second practice in the adaptive scheduling block discusses additional, ad-hoc alertness breaks, taken as-needed, during a typical shift. We generally refer to both impromptu alertness breaks and invited alertness breaks as a *smart breaks* practice. The use of smart breaks prevents extended periods of time on task, punctuating periods of monotony. As mentioned in the introduction, where the notion of fatigue is discussed, a substantial body of research has observed that the interruption of monotonous tasks through both physical and cognitive stimulation can reduce the risks induced by hypovigilance [81]. Impromptu alertness breaks are self-initiated by autonomous specialists, for example based on an indication of high fatigue from a PFS (as in the example depicted in Figure 10). Invited alertness breaks, conversely, are offered to the autonomous specialist and may be triggered in several ways, for example from human or automated fatigue observations, or upon patterns of poor ICT performance. Invited alertness breaks can also occur automatically and they may help specialists act upon fatigue that may have otherwise gone unnoticed. This is because fatigue can, at times, be difficult to self-identify. Invited breaks are one example of a way to provide timely support to a fatigued specialist.

The third methodology within the adaptive scheduling block is the use of auxiliary tasks assignments. The *auxiliary tasks assignments* process is a workforce management practice that employers could use to assign fatigued specialists to non-safety-critical off-road tasks as operational needs dictate. This can help the specialist feel productive and appreciated for their openness and proactive fatigue communication. As mentioned before, flexibility and availability to re-staff a shift with a different autonomous specialist, while redirecting the fatigued individual to a different role, are of primary importance to encourage alertness and, at the same time, ensure openness and transparency of reporting without fear of retaliation.

## OVERVIEW AND MAPPING

The previous section delved into the various methodologies that comprise Waymo's proposed FRM framework, which rests on the solid foundation of clear recommendations and guidelines. Earlier in this report, we highlighted how Waymo's framework for fatigue risk management accounts for the three different layers of prevention, monitoring, and mitigation (typical of defense-in-depth approaches [83]) - each applied across the entire lifecycle of operations: before driving, while in the driver's seat, and after driving. This concept, which was presented in Figure 5 (here reproposed in Figure 18, for ease of reference), is actually embedded within the methodologies presented in this report through a precise mapping across all the building blocks of Figure 6.

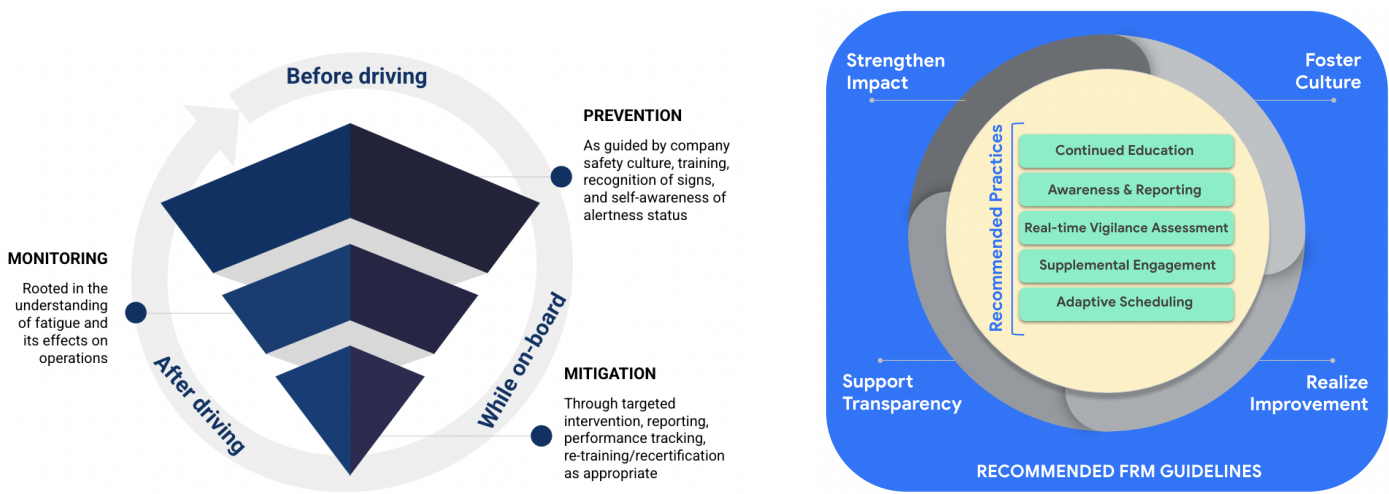

Figure 18. A layered approach to fatigue risk management, and the building block of our recommended FRM framework (juxtaposition of prev. Figure 5 and Figure 6 for reference)

Such mapping leads to Figure 19, which details how the methodologies and implementation blocks covered in this section translate, in practice, to twenty practices that span the lifecycle of autonomous vehicle testing operations with autonomous specialists before, during, and after each on-road session. Many of the recommended methodologies listed in Figure 19 are targeting prevention and mitigation strategies that address fatigue-induced risks before getting into the vehicle itself. Our experience, and the substantial body of research that exists on the risks induced by fatigue, as indicated that such a multi-pronged approach is necessary (see, for example, [24; 82]). Different reasons can lead to fatigue onset, and no single solution to date proved sufficiently effective to tackle these risks to an acceptable degree.

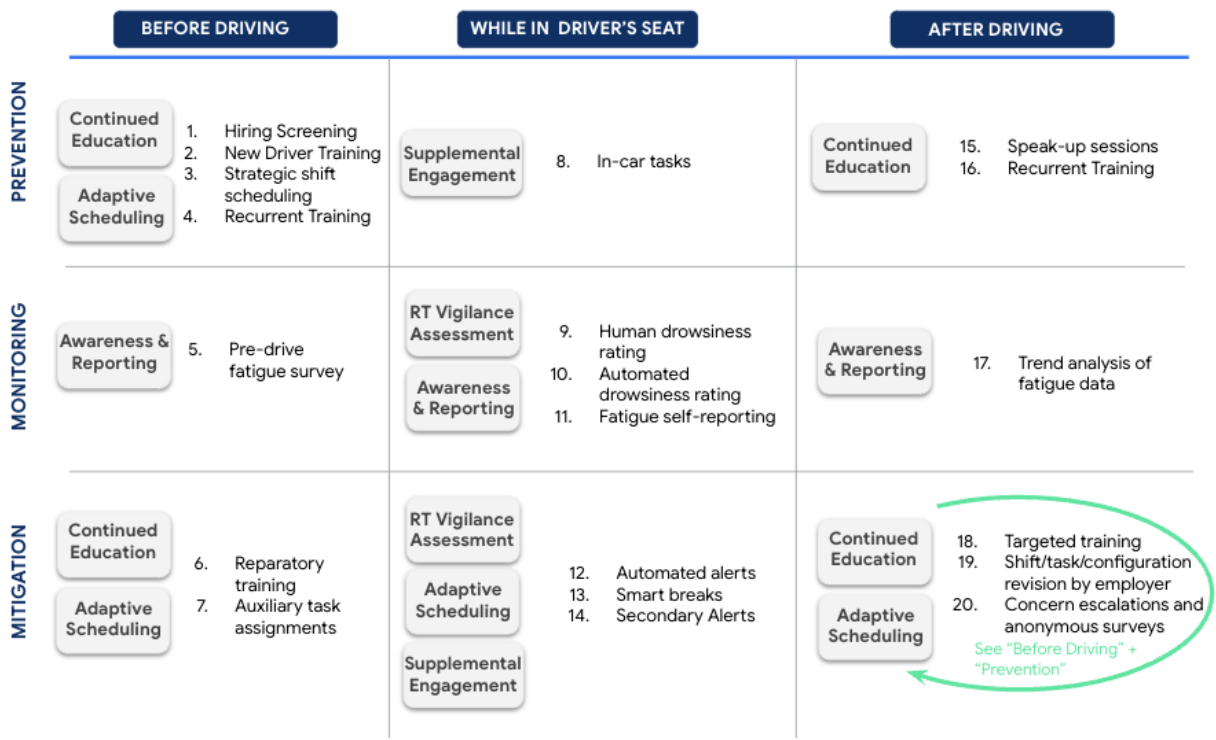

Figure 19. Mapping of recommended elements, pillars, and methodologies for FRM

As highlighted in the previous section, recent literature discussed limitations associated with the measurement of effectiveness of some of the methodologies listed in Figure 19 [75; 78; 84]. However, the majority of the literature addresses tasks other than advanced automation monitoring, and more research is needed to fully validate accuracy and effectiveness of multiple methodologies in the context of duties such as those carried out by autonomous specialists. That is why we recommend an approach that uses multiple indicators and metrics, providing measurement overlap and redundancies, to ensure appropriate coverage across multiple lines of defenses, and along all layers recognized by the literature on fatigue-risk trajectories [82; 85]. While this is a starting point, our experience will help us continuously improve on it. We remain committed to continue researching and identifying the best solutions to address fatigue, and to investigate new advancements of external research, as those become available.

Finally, the FRM framework is more than the sum of its parts. The recommended practices covered in this report are truly indicative of the full-circle visualized in Figure 6, where culture, performance, transparency, and impact must progress hand in hand to ensure the safety and robustness of an approach that is applied every day on the road.

# 3. Conclusions

Waymo's mission to reduce traffic injuries and fatalities on public roads rests on the safe and responsible deployment of the Waymo Driver, a Level 4 automated driving system (ADS). Autonomous specialists are central to the safe development and testing of the Waymo Driver. Yet, our early testing and a substantial body of literature indicated that any human — even those trained and specifically told to continuously monitor the operation of the ADS — can be subject to complacency and an altered state of attention (i.e., distracted, drowsy, or otherwise impaired) that can hinder the safety of the overall endeavor.

This report thus provided an in-depth look at a portfolio of recommended methodologies organized in a systematic Fatigue Risk Management framework that Waymo is proposing for counteracting fatigue-induced risks associated with autonomous specialists' operation of autonomous vehicles.

Waymo's FRM framework recommends a wide range of fatigue countermeasures to ensure that autonomous specialists can effectively oversee automated driving systems during testing operations. Autonomous specialists play a central role during ADS development, which include operations tied to generating maps, undergoing durability testing, testing new hardware, and testing early versions of new software. Although the methodologies outlined here regarding the Fatigue Risk Management framework are intended to address fatigue-induced impairment, a form of insufficient attention (see Figure 1), they may also positively affect other forms of inattention and complacency, which were not covered in detail in this report.

As we deploy the Waymo Driver in new operating environments, we continue to learn from our experience on the road and we strive to ensure that our FRM program remains highly effective in those new domains. This report presents recommendations derived from continuously advancing research, and the methodologies here proposed are not to be intended as part of a static framework, but rather as a living and constantly evolving commitment to always prioritize the safety of our riders, team, and all who share the road with us.

A last important consideration stems from the application of Waymo's FRM framework to both single-specialist and dual-specialist testing configurations. Figure 8 provided an example in which autonomous specialists undergoing training commence operations in dual-specialist mode, and later transition to the single configuration. While dual-specialist programs may prove to be effective in mitigating fatigue in some instances, dual specialist programs should not be relied upon exclusively for fatigue detection and thus cannot substitute a systematic FRM program due to a number of reasons, ranging from inadequate positioning for spotting fatigue signs, to risks of under-reporting due to peer pressure, and the fact that dual specialists would also be subject to fatigue.

Finally, in sharing our framework for fatigue risk management, we hope to inform the industry best practices for fatigue monitoring and their application in real-life systems. Our past experience has shaped Waymo's decision to tackle Level 4 automation. That experience has also helped us craft an FRM framework that we are today proposing to the industry through publication: we hope that our learnings can prove useful in lessening the risks associated with complacency, overreliance, and misuse of technology associated with lower levels of automation as well. While we highlighted many existing practices that this proposal draws and builds upon, it is our hope that the details shared in this report can foster the continued dialogue with others in the industry toward achieving the full potential promised by fully autonomous technology, while carefully managing risks along the way.

Definition of Terms and Concepts. *Surface Vehicle Information Report. SAE, USA.*

[25] Engström J, Monk CA, Hanowski RJ, Horrey WJ, Lee JD, McGehee DV, Regan M, Stevens A, Traube E, Tuukkanen M, Victor T, Yang CYD. A conceptual framework and taxonomy for understanding and categorizing driver inattention. Brussels, Belgium: European Commission; 2013.

[26] National Center for Statistics and Analysis. (2017, October). Drowsy Driving 2015 (Crash•Stats Brief Statistical Summary. Report No. DOT HS 812 446). Washington, DC: National Highway Traffic Safety Administration.

[27] Klauer, S.G., Dingus, T.A., Neale, V.L., Sudweeks, J.D. and Ramsey, D.J., 2006. The impact of driver inattention on near-crash/crash risk: An analysis using the 100-car naturalistic driving study data.

[28] Tefft, B.C., 2012. Prevalence of motor vehicle crashes involving drowsy drivers, United States, 1999–2008. *Accident Analysis & Prevention*, *45*, pp.180-186.

[29] Akerstedt, T., Arendt, J., Cassel, W., Dinges, D., Englund, L., Findley, L., Folkard, S., George, C., Gillberg, M., Guilleminault, C., Hack, M., Haraldsson, P-O., Hartley, L., Hetta, J., Horne, J., Kecklund, G., Krieger, J., Landstrom, U., Nicholson, A., Pack, A., Parkes, D., Partinen, M., Philip, P., Reyner, L., Rosekind, M., Samel, A., Spencer, M., and Zulley, J. (2000). Consensus statement: Fatigue and accidents in transport operations. Journal of Sleep Research, 9, 395

[30] Rosekind, M.R., Gander, P.H., Miller, D.L., Gregory, K.B., Smith, R.M., Weldon, K.J., Co, E.L., Mcnally, K.L. and Lebacqz, J.V., 1994. Fatigue in operational settings: examples from the aviation environment. Human factors, 36(2), pp.327-338.

[31] Caldwell, J.A., 2005. Fatigue in aviation. Travel medicine and infectious disease, 3(2), pp.85-96.

[32] Goode, J.H., 2003. Are pilots at risk of accidents due to fatigue?. Journal of safety research, 34(3), pp.309-313.

[33] Rosekind, M.R., Boyd, J.N., Gregory, K.B., Glotzbach, S.F. and Blank, R.C., 2002. Alertness management in 24/7 settings: lessons from aviation. Occupational Medicine-Philadelphia-, 17(2), pp.247-260.

[34] Waterhouse, J., Edwards, B., Atkinson, G., Reilly, T., Spencer, M. and Elsey, A., 2016. Occupational factors in pilot mental health: sleep loss, jet lag, and shift work. Hubbard, T. and
41